\documentclass[runningheads]{llncs}

\usepackage{eccv}

\usepackage{eccvabbrv}

\usepackage{graphicx}
\usepackage{booktabs}
\usepackage{caption}
\usepackage{array}
\usepackage{multirow}
\usepackage{amsmath}
\usepackage{comment}
\usepackage{array}
\usepackage{multicol}
\usepackage{mathrsfs}
\usepackage{makecell}
\usepackage{subcaption}

\usepackage{caption} % for \captionof

\usepackage{amssymb}  % for \checkmark
\usepackage{pifont}   % for \xmark

\newcommand{\xmark}{\ding{55}}  % define \xmark if not already defined

\usepackage[accsupp]{axessibility}  % Improves PDF readability for those with disabilities.

\usepackage{hyperref}

\usepackage{orcidlink}

\begin{document}

% ---------------------------------------------------------------
% TODO REVIEW: Replace with your title
% \title{Supplementary Material for\\``SIGNER: Temporally Grounded Sign Language Generation via Time-Resolved Conditioning''}
\title{SIGNER: Temporally Grounded Sign Language Generation via Time-Resolved Conditioning}
% Temporally Grounded Sign Language Generation via Time-Resolved Conditioning

% TODO REVIEW: If the paper title is too long for the running head, you can set
% an abbreviated paper title here. If not, comment out.
\titlerunning{SIGNER}

% TODO FINAL: Replace with your author list. 
% Include the authors' OCRID for the camera-ready version, if at all possible.
% \author{Taeryung Lee\inst{1}\orcidlink{0000-1111-2222-3333} \and
% Hyeongjin Nam\inst{2}\orcidlink{1111-2222-3333-4444} \and
% Gyeongsik Moon\inst{3}\orcidlink{2222--3333-4444-5555} \and
% Kyoung Mu Lee\inst{2}\orcidlink{1111-2222-3333-4444} 
% }

\author{
Taeryung Lee\inst{1}\orcidlink{0009-0005-5585-1413} \and
Hyeongjin Nam\inst{2}\orcidlink{0009-0004-9387-407X} \and
Gyeongsik Moon\inst{3}$^\dag$\orcidlink{0000-0003-0610-7936} \and \\
Kyoung Mu Lee\inst{1,2}$^\dag$\orcidlink{0000-0001-7210-1036}
}

\institute{
\textsuperscript{1}IPAI, 
\textsuperscript{2}Dept. of ECE \& ASRI, Seoul National University
\\
\textsuperscript{3}Dept. of CSE, Korea University
}

\begingroup
\renewcommand{\thefootnote}{}
\footnotetext{\textsuperscript{\dag} Corresponding authors.}
\endgroup

% TODO FINAL: Replace with an abbreviated list of authors.
\authorrunning{T.~Lee et al.}
% First names are abbreviated in the running head.
% If there are more than two authors, 'et al.' is used.

% TODO FINAL: Replace with your institution list.
% \institute{IPAI, \and Dept. of ECE \& ASRI, Seoul National University, Korea \and
% Dept. of CSE, Korea University, Korea}
\maketitle

% \begin{abstract}
%   The abstract should concisely summarize the contents of the paper. 
%   While there is no fixed length restriction for the abstract, it is recommended to limit your abstract to approximately 150 words.
%   Please include keywords as in the example below. 
%   This is required for papers in LNCS proceedings.
%   \keywords{First keyword \and Second keyword \and Third keyword}
% \end{abstract}

% \maketitle

% Goal: Temporal grounding: temporal correspondence between glosses and their realized sign segments
% Previous methods: Global fusion (e.g. cross attention, our LTF와 대비)
% Methodology: Time-Resolved Conditioning = (Local temporal fusion + temporal gloss condition)
% Condition: Temporal Gloss Condition
% Method: Local Temporal Fusion (LTF) = “temporally localized conditioning을 구현하는 모듈”

\begin{abstract}
Sign language generation (SLG), also known as text-to-sign generation, aims to bridge the communication gap between signers and non-signers.
Unlike many other generative tasks, SLG must satisfy two fundamental linguistic constraints. 
First, sign language expresses meaning through a sequence of gestures aligned with word-like units called glosses, and therefore requires correct lexical ordering to preserve intended meaning. 
Second, each gesture should faithfully reflect the intended gloss (semantic accuracy).
Despite recent progress, existing SLG methods frequently produce signs with incorrect lexical order and low semantic accuracy.
A common limitation of prior approaches stems from globally fused conditioning strategies, which weaken \textit{temporal grounding}, the temporal correspondence between glosses and their realized sign segments.
This often leads to incorrect lexical order and semantically ambiguous signs.
To address this limitation, we propose \textbf{SIGNER}, a \textbf{SIGN} language generation framework with tim\textbf{E}-\textbf{R}esolved conditioning to ensure temporal grounding, leveraging a temporal-gloss condition and local temporal fusion (LTF).
SIGNER constructs a temporal-gloss condition by estimating a gloss sequence and its durations from input text, and assigning gloss semantics across the temporal dimension. 
We then introduce LTF, a temporally grounded fusion module that integrates the temporal-gloss condition within a constrained temporal window during denoising.
By enforcing temporal locality in condition fusion, LTF preserves temporal grounding, leading to correct lexical ordering and clearer per-gloss semantics.
Experiments on Phoenix-2014T and CSL-Daily demonstrate state-of-the-art performance, further supported by motion-smoothness analysis.
% We then introduce Local Temporal Fusion (LTF), a temporal-locality-preserving fusion module which integrates the temporal-gloss condition within a constrained temporal window during denoising.
The project page is available \href{https://taeryunglee.github.io/projects/signer/}{here}.


\end{abstract}

\clearpage

\begin{figure}[t]
    \hspace{-3mm}
    \includegraphics[width=\linewidth]{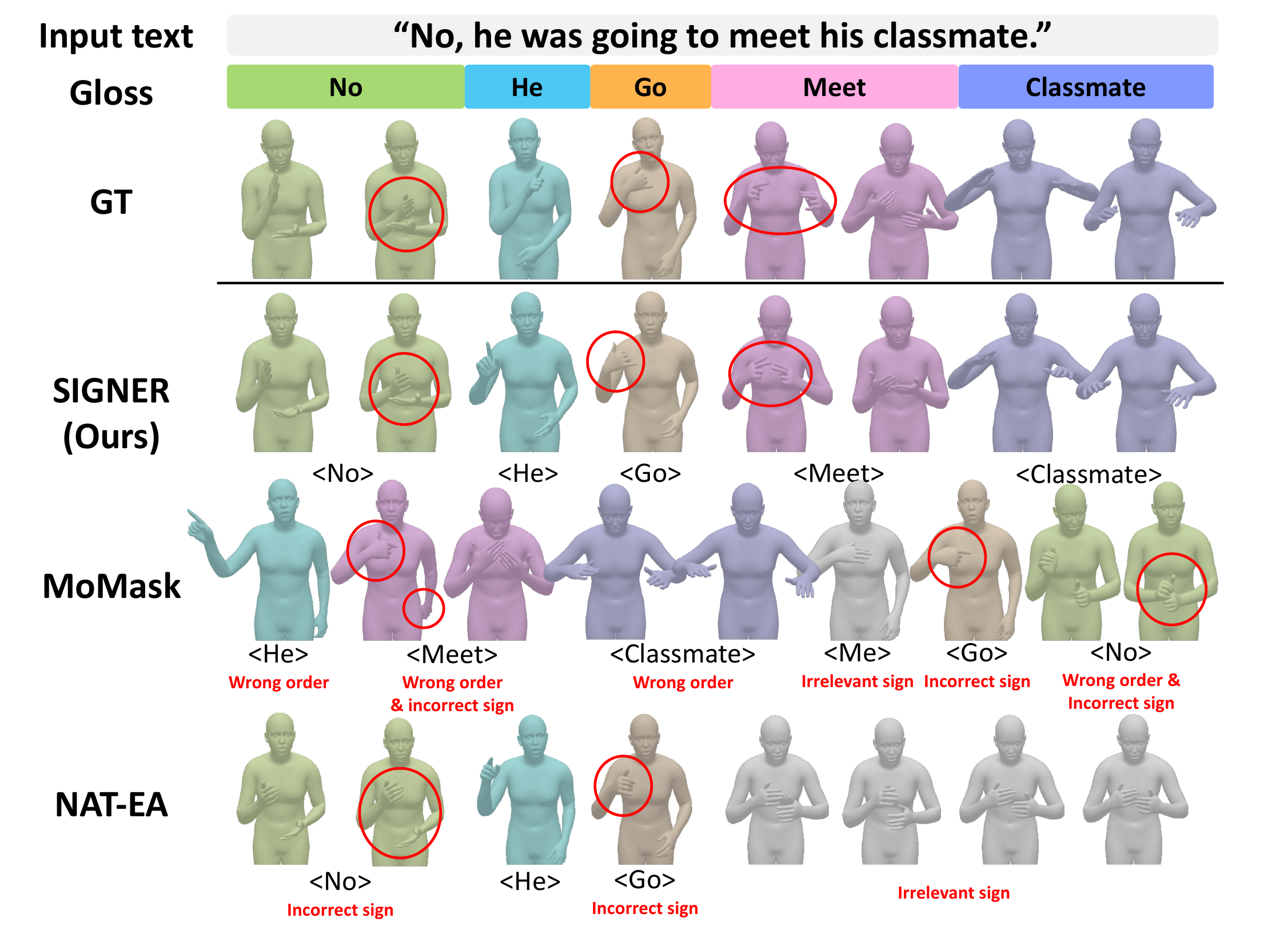}
    \caption{
        \textbf{Qualitative comparison.}
        We compare the generated signs from our \textbf{SIGNER}~and prior methods, MoMask~\cite{momask} and NAT-EA~\cite{natat}, all trained on CSL-Daily.
        We represent the glosses with colors, and the sign segments corresponding to each gloss are painted in the same color.
        Human-interpreted glosses of generated signs are denoted in brackets below the signs.
        While the signs generated by the prior methods exhibit incorrect lexical order and include inaccurate signs, our SIGNER produces accurate signs in the correct order.}
    \label{fig:teaser}
\end{figure}

\section{Introduction}~\label{sec:intro}

\begin{figure}[t]
    % \hspace{-5mm}
    \includegraphics[width=\linewidth]{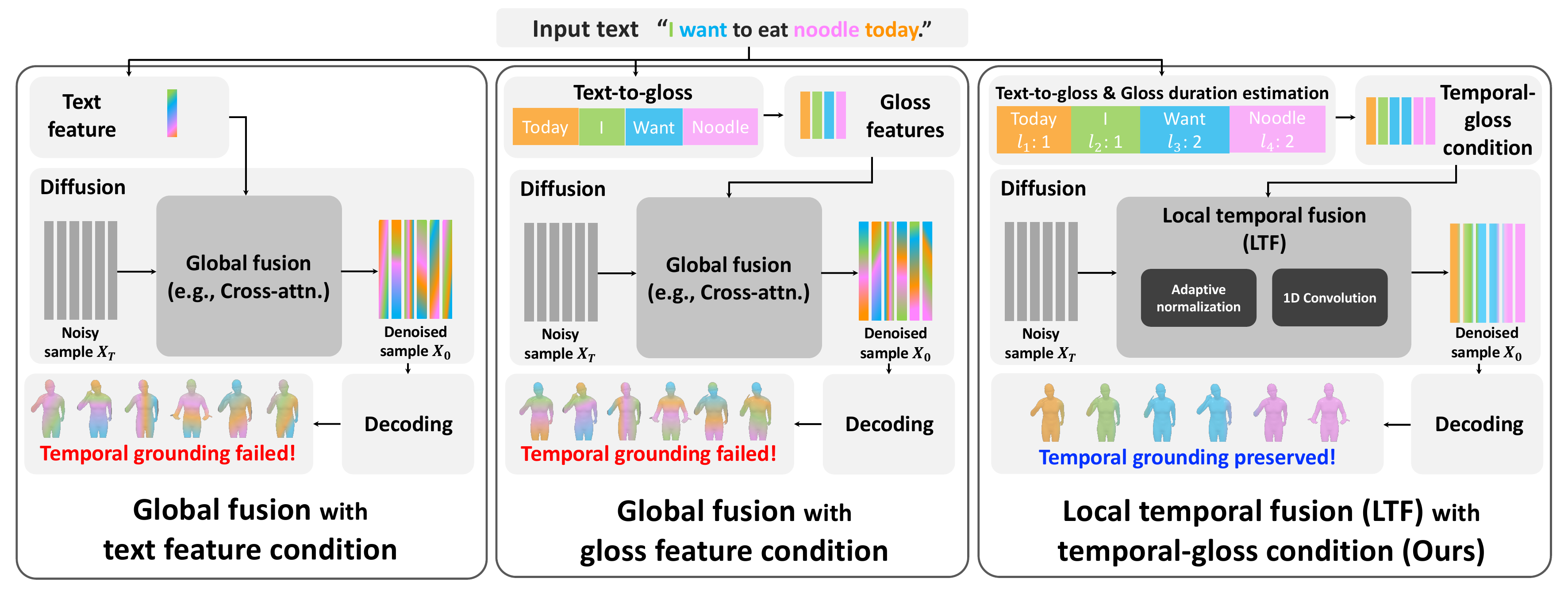}
    \caption{\textbf{Temporally grounded sign language generation.}
    Previous SLG approaches often rely on global fusion (e.g., cross-attention), which integrates text/gloss features over the entire sequence (left/middle), without ensuring temporal grounding (i.e., temporal gloss-sign correspondence).
    In contrast, our method realizes time-resolved conditioning by combining temporal-gloss condition with local temporal fusion (LTF) to preserve temporal grounding (right).}
    \label{fig:comparison}
\end{figure}

Sign language is a primary communication medium for millions of deaf and hard-of-hearing (DHH) individuals worldwide.
However, communication between signers and non-signers often depends on expert interpreters.
Text-to-sign language generation (SLG)~\cite{soke, nsa, spoken2sign, t2s-gpt} is a key step toward lowering this barrier, enabling applications such as automated subtitles and virtual assistants.

Unlike many other generative tasks such as text-to-image~\cite{stablediffusion}, SLG must satisfy strict linguistic constraints. 
Sign language expresses meaning through a sequence of gestures aligned with word-like linguistic units called glosses. 
Each gloss corresponds to a sign segment, and these segments must appear in the correct \textit{lexical order} to preserve the intended meaning~\cite{napoli2014order, cheng2019acquiring}.
% Moreover, the lexical semantics of individual glosses must remain temporally distinct across sign segments.
% Moreover, semantic accuracy is equally important, which means, how faithfully each gesture reflects the intended gloss.
In addition to lexical order, \textit{semantic accuracy}—how faithfully each gesture reflects the intended gloss—is equally crucial for preserving clarity in sign language communication.
These linguistic properties imply that SLG requires explicit \textit{temporal grounding}, namely temporal correspondence between gloss semantics and their realized sign segments (illustrated in Figure~\ref{fig:teaser} by color correspondence).

Despite recent advances, existing SLG methods frequently fail to generate semantically accurate signs in correct lexical order, as illustrated in Figure~\ref{fig:teaser}.
A common limitation among prior works is the use of global fusion~\cite{mdm, momask, soke, stablediffusion}, where conditioning signals are shared across the entire sequence (Figure~\ref{fig:comparison}, left).
Such global fusion provides no explicit indication of which gloss should be presented at a particular timestep, often resulting in incorrect lexical order and semantically ambiguous signs.
Some works incorporate gloss sequences to better reflect lexical structure~\cite{g2p-ddm, natat} (Figure~\ref{fig:comparison}, middle).
However, even in these methods, gloss-based conditions are still globally fused, which limits explicit preservation of temporal grounding and leaves the limitation of global fusion only partially addressed (Figure~\ref{fig:teaser}, last row).

% Restricting fusion to a local temporal window, LTF prevents non-target gloss semantics from leaking across time into the current timestep.
% This prevents cross-gloss competition that causes lexical-order errors, and it prevents boundary blurring between adjacent gloss segments, improving semantic accuracy.

% To address these structural limitations of globally fused conditioning, we propose \textbf{SIGNER}, a \textbf{SIGN} language generation framework by tim\textbf{E}-\textbf{R}esolved conditioing as illustrated in Figure~\ref{fig:comparison}, right.
% SIGNER aims to preserve temporal grounding through time-resolved conditioning that restricts condition fusion to local temporal windows, thereby mitigating temporal ambiguity that often leads to incorrect lexical ordering and semantically wrong gestures in prior approaches.

To address these limitations, we propose \textbf{SIGNER}, a \textbf{SIGN} language generation framework with tim\textbf{E}-\textbf{R}esolved conditioning (Figure~\ref{fig:comparison}, right).
In contrast to prior methods based on global fusion, which share conditioning signals across the entire sequence, SIGNER preserves the temporal correspondence between glosses and their realized sign segments (\textit{i.e.,} temporal grounding).
By explicitly maintaining temporal grounding during generation, SIGNER prevents gloss information from being mixed across time, which often causes incorrect lexical order and semantically ambiguous signs.

% To provide temporally indexed conditioning signals, SIGNER constructs time-resolved conditions from the input text (bottom-left).
% Specifically, we obtain a gloss sequence from the off-the-shelf text-to-gloss model, predict gloss durations with a learned estimator, and expand gloss embeddings along the temporal dimension to construct time-resolved conditions.
% This construction encodes lexical order and distributes gloss semantics along the temporal axis, yielding temporally grounded conditioning for generation.

% Our key component is Local Temporal Fusion (LTF), which integrates time-resolved conditions into diffusion denoising while restricting fusion to a local temporal window (right).
% This localized fusion implements temporally localized conditioning to prevent global mixing of gloss information across time and preserve temporal grounding, enabling correct lexical ordering with semantically distinguishable sign segments.
% Specifically, LTF combines adaptive normalization (scale/shift modulation) with a temporal 1D convolution for local context aggregation, resulting in temporally coherent and semantically precise sign generation with smooth and natural transitions between adjacent lexical segments.

SIGNER implements time-resolved conditioning by combining a temporal-gloss condition with local temporal fusion (LTF).
% We construct the temporal-gloss condition by deriving a gloss sequence from the input text, predicting gloss durations, and expanding gloss embeddings along the temporal axis.
We construct the temporal-gloss condition by deriving a gloss sequence from the input text, estimating gloss durations, and repeating each gloss embedding for its estimated duration to form a time-resolved sequence.
Our key contribution is LTF, which integrates this condition into diffusion denoising within a local temporal window.
In contrast to global fusion that integrates the condition globally across time and weakens time-specific guidance, LTF preserves temporal grounding by applying the condition locally, which enforces the correct lexical order.
Moreover, this locality constraint prevents semantic overmixing across adjacent segments, improving semantic accuracy.

We evaluate SIGNER on Phoenix-2014T~\cite{phoenix} and CSL-Daily~\cite{csldaily}, where it substantially outperforms prior SLG methods on back-translation-based metrics.
Beyond automatic evaluation, we additionally conduct a motion-smoothness analysis (e.g., peak jerk statistics) to validate overall motion quality.
Our contributions are summarized as follows.
\begin{itemize}
    % \item We propose \textbf{SIGNER}, an SLG framework based on time-resolved condition with temporally localized conditioning to generate accurate sign language in correct lexical order.
    % \item We introduce Local Temporal Fusion (LTF), a locality-preserving fusion mechanism that constrains conditioning fusion within a local temporal window, explicitly enforcing temporal grounding during generation.    
    % \item We conduct comprehensive evaluations on two benchmarks, demonstrating that temporally localized conditioning consistently improves lexical ordering, semantic clarity, and overall generation quality.
    \item We propose \textbf{SIGNER}, a sign language generation framework via time-resolved conditioning to enforce lexical order of generated signs.
    \item We introduce local temporal fusion (LTF), which integrates the temporal-gloss condition locally during denoising to ensure the temporal grounding.
    \item LTF also enhances semantic accuracy by mitigating overmixing of gloss semantics across time.
\end{itemize}

\section{Related works}

\noindent\textbf{Sign language generation (SLG).}
SLG synthesizes sign gesture sequences from spoken-language text, enabling machine-mediated communication for deaf and hard-of-hearing individuals~\cite{nsa,progressivetransformers,adversarialtraining,nar,spoken2sign,soke,signavatars,t2s-gpt}.
Recent methods span diverse generation paradigms, including diffusion-based generation~\cite{nsa}, autoregressive token generation~\cite{soke,t2s-gpt}, and retrieval-and-concatenation pipelines~\cite{spoken2sign}.
Despite progress, two limitations remain central: generating signs in correct lexical order and maintaining semantic accuracy at the segment level.
A common cause is global fusion, where text or gloss features are shared across the entire sequence, weakening temporal grounding and leading to cross-time confusion of lexical cues and temporally mixed semantics.
% Several works incorporate gloss features~\cite{g2p-ddm} or temporal-gloss conditions~\cite{natat} to better reflect lexical structure, but gloss conditions are still typically fused globally via cross attention, leaving these limitations only partially addressed.
Several works incorporate gloss features~\cite{g2p-ddm} or temporal-gloss conditions~\cite{natat} to better reflect lexical structure.
However, such conditions are still globally fused into generation process via cross-attention, leaving these limitations not fully resolved.
% Retrieval-based pipelines such as Spoken2Sign~\cite{spoken2sign} can encourage lexical order by construction, yet often struggle with realistic transitions due to limited modeling of fine-grained temporal dependencies.
Retrieval-based pipelines such as Spoken2Sign~\cite{spoken2sign} can enforce lexical order by sequentially connecting retrieved sign segments, but they often yield unnatural transitions due to limited modeling of temporal dependencies across segment boundaries.
In contrast, our approach explicitly enforces temporal grounding through time-resolved conditioning, combining a temporal-gloss condition with LTF.
This directly addresses the limitations of previous approaches using global fusion, improving lexical order and semantic accuracy.

\noindent\textbf{Human motion generation.}
Human motion generation aims to synthesize temporally coherent body movements from text descriptions~\cite{mdm,mld,t2mgpt,MultiAct,t2lm,motiongpt,momask,petrovich21actor,teach,st2m,priormdm,chuan2022tm2t,petrovich22temos,zhang2024motiondiffuse,wu2025mg}.
ACTOR~\cite{petrovich21actor} learns action-conditioned motion embeddings with a Transformer VAE.
MDM~\cite{mdm} applies diffusion in motion space for text-to-motion generation.
MotionGPT~\cite{motiongpt} models motion as discrete tokens for unified pretraining.
MoMask~\cite{momask} uses masked modeling over a hierarchical VQ representation.
While these methods can be adapted to SLG, sign language imposes additional linguistic constraints such as lexical order and gloss-level semantic specificity.
Our temporally-localized conditioning is designed to better respect these constraints during generation.

\begin{figure*}[t]
    \centering
    \includegraphics[width=\linewidth]{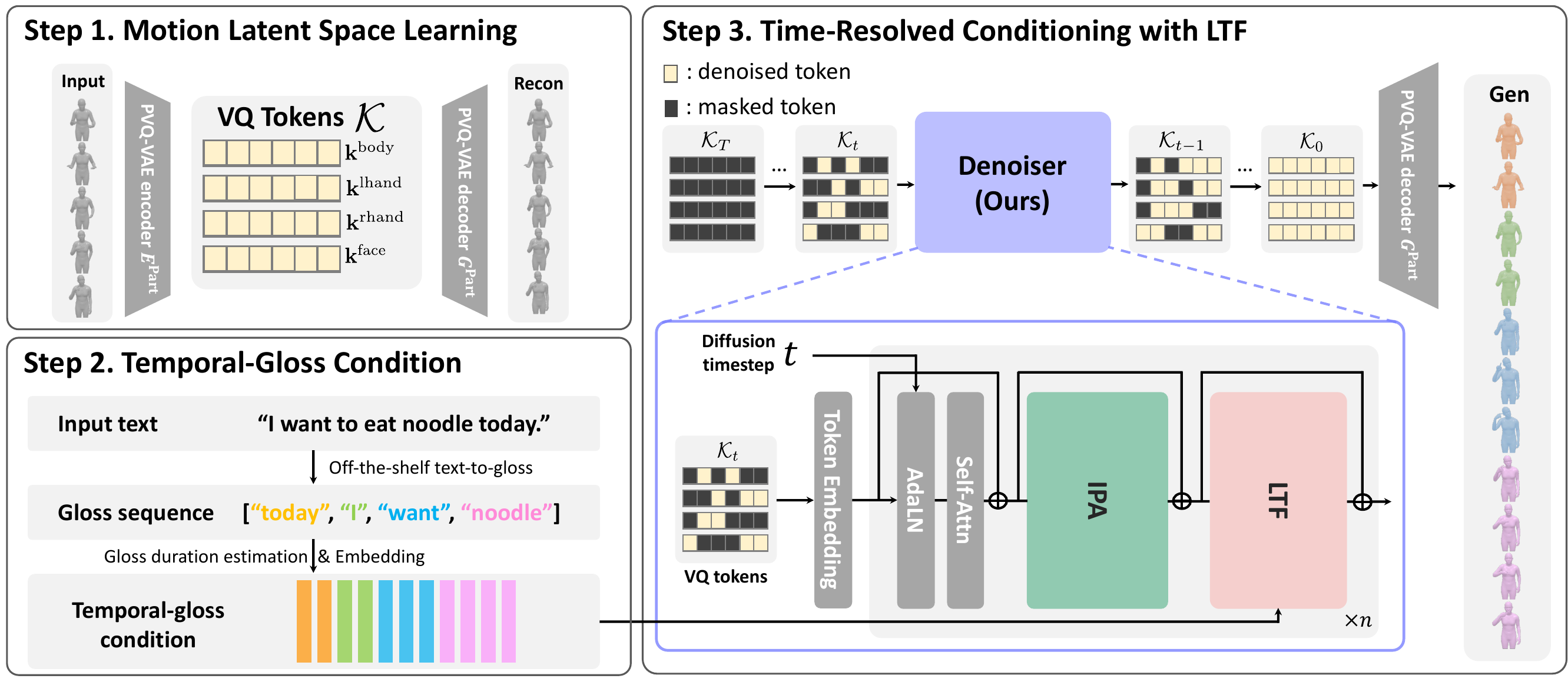}
    \caption{\textbf{Overview.} 
    Our method aims to model temporal correspondence between the glosses and their associated sign segments (temporal grounding).
    We first learn a motion latent space using PVQ-VAE (top-left).
    Next, we construct temporal-gloss condition from the input text (bottom-left), providing time-resolved conditioning signals.
    Finally, SIGNER generates sign language sequences by integrating temporal-gloss conditions within localized temporal windows via local temporal fusion (LTF), enabling accurate temporal grounding.
    }
    % \vspace{-3mm}
    \label{fig:overview}
\end{figure*}
\section{Preliminary}
% In this section, we provide the background of our architecture on VQ-VAE~\cite{vqvae} and discrete diffusion model~\cite{vqdiffusion}.

\noindent\textbf{Motion Representation.}
We utilize the human motion representation defined on 43 joints of upper-body and hands, consisting of 3D joint position, 6D rotation and 3D velocity. 
We also leverage facial expression parameters and jaw poses. 
Therefore, our pose representation consists of $43 \times (3 + 6 + 3) + 10 + 3 = 529$ dimensions.

% \subsection{Learning Discrete Latent Space}~\label{subsec:preliminary_latent}
\noindent\textbf{VQ-VAE.}\label{subsec:preliminary_latent} VQ-VAE~\cite{vqvae, t2mgpt} comprises an encoder $E$, a decoder $G$, and a codebook $\mathcal{Z} = \{\mathbf{z}_k \in \mathbb{R}^{d}\}_{k=1}^K$, where $K$ is the number of codebook entries and $d$ is the dimensionality of each code vector.
Given an input motion $\mathbf{X} \in \mathbb{R}^{L \times D}$, where $L$ and $D$ denote the sequence length and feature dimension, respectively, the encoder maps it to a latent representation $\mathbf{Z}^e = E(\mathbf{X}) \in \mathbb{R}^{L' \times d}$.
The length of the latent sequence is defined as $L' = \lfloor L/r \rfloor$, where $r$ denotes the temporal reduction rate of the encoder $E$.
We denote the encoder outputs and quantized latent representation as $\mathbf{Z}^e$ and $\mathbf{Z}^q$, respectively.
Each latent vector $\mathbf{Z}^e_i$ at temporal step $i$ is quantized to the nearest code $\mathbf{z}_k$:
\begin{equation}
\mathbf{Z}^q_i = \mathbf{z}_k, \quad \text{where } k = \arg\min_{k'} \|\mathbf{Z}^e_i - \mathbf{z}_{k'}\|_2.
\end{equation}
The decoder reconstructs the motion from the quantized latent codes $\hat{\mathbf{X}} = G(\mathbf{Z}^q)$.
The VQ-VAE is trained to minimize the following loss function:
\begin{equation}\label{eq:vqvae_loss}
\mathcal{L}_\text{VQ} = \|\mathbf{X} - \hat{\mathbf{X}}\|_2^2 + \|\text{sg}(\mathbf{Z}^e) - \mathbf{Z}^q\|_2^2 + \beta \|\mathbf{Z}^e - \text{sg}(\mathbf{Z}^q)\|_2^2,
\end{equation}
where $\text{sg}(\cdot)$ is the stop-gradient operation, and $\beta$ is a loss weight.

% \begin{figure*}[t]
%     \centering
%     \includegraphics[width=0.9\linewidth]{fig/overview.pdf}
%     \caption{\textbf{Overview.} We first construct the motion latent space with PVQ-VAE (left). In generation phase (right), we first encode input text into gloss-level condition with off-the-shelf text-to-gloss model~\cite{chen2022twostream} and length sampling. Our proposed denoiser with TAC generates the tokens in the motion latent space and forwards them into the PVQ-VAE decoder to generate sign language.}
%     % \vspace{-5mm}
%     \label{fig:overview}
% \end{figure*}

% \subsection{Vector-Quantized Diffusion Model}
\noindent\textbf{Discrete diffusion.} Discrete diffusion~\cite{vqdiffusion} operates over discrete tokens, a finite set of codebook indices, obtained from a pre-trained VQ-VAE. 
Unlike continuous diffusion models~\cite{ddpm}, which add Gaussian noise to real-valued inputs, this approach defines a Markov diffusion process over discrete tokens.
% a finite set of codebook indices.

% \textbf{Forward process.}
Let $\mathbf{k}_0 \in \{1, \dots, K\}^{L'}$ denote the original sequence of discrete tokens from the codebook. 
The forward diffusion process gradually corrupts $\mathbf{k}_0$ over $T$ timesteps by randomly replacing tokens, defined by a transition matrix $Q_t$:
\begin{equation}
q(\mathbf{k}_t \mid \mathbf{k}_{t-1}) = \mathbf{v}^\top(\mathbf{k}_t) Q_t \mathbf{v}(\mathbf{k}_{t-1}),
\end{equation}
where $\mathbf{v}(\cdot)$ denotes the one-hot representation of a discrete token, and $Q_t \in \mathbb{R}^{K \times K}$ is a stochastic matrix with rows summing to 1. 
The transition matrix is constructed using hyperparameters $\alpha_t$, $\beta_t$, and $\gamma_t$, which control the probabilities of token retention, replacement with random tokens, and masking, respectively.
At the final timestep $T$, all tokens are fully masked, and the masked sequence serves as the starting point for the reverse denoising process, analogous to sampling from a standard Gaussian distribution in continuous DDPM frameworks.

The goal of the reverse process is to recover $\mathbf{k}_0$ from a noisy $\mathbf{k}_t$. This is done by estimating the reverse transition:
\begin{equation}
p_\theta(\mathbf{k}_{t-1} \mid \mathbf{k}_t, \mathbf{y}) = \sum_{\tilde{\mathbf{k}}_0} q(\mathbf{k}_{t-1} \mid \mathbf{k}_t, \tilde{\mathbf{k}}_0) \, p_\theta(\tilde{\mathbf{k}}_0 \mid \mathbf{k}_t, \mathbf{y}),
\end{equation}
where $\tilde{\mathbf{k}}_0$ is the predicted denoised discrete token sequence, and $\mathbf{y}$ is a conditional input (e.g., text prompt). 
The denoiser model, which can be designed for the specific purpose such as sign language generation, predicts the initial state $p_\theta(\tilde{\mathbf{k}}_0 \mid \mathbf{k}_t, \mathbf{y})$.
Denoiser is trained to estimate the posterior transition distribution $q(\mathbf{k}_{t-1} \mid \mathbf{k}_t, {\mathbf{k}}_0)$ with the variational lower bound~\cite{vqdiffusion}.

% \subsection{Human Motion Representation}~\label{subsec:representation}
% Inspired from HumanML3D~\cite{t2m} and HumanTOMATO~\cite{humantomato}, we design a whole-body motion representation for sign language.
% Our representation includes the upper body, hands, and facial expressions, which are the primary articulators in sign communication. 
% We assume that root joints are fixed to the origin.
% Each pose $\mathbf{X}_i$ at timestep $i$ consists of four components: local joint positions $\mathbf{p}_i \in \mathbb{R}^{3J}$ for $J = 43$ upper-body and hand joints; local joint rotations $\mathbf{r}_i \in \mathbb{R}^{6J}$ using a continuous 6D representation~\cite{zhou2019rotation6d}; local joint velocities $\mathbf{v}_i \in \mathbb{R}^{3J}$ obtained via finite differences; jaw pose and low-dimensional facial expression features $\mathbf{f}_i \in \mathbb{R}^{13}$.
% Thus, each pose vector is represented as $\mathbf{X}_i = [\mathbf{p}_i; \mathbf{r}_i; \mathbf{v}_i; \mathbf{f}_i] \in \mathbb{R}^{(3+6+3)J + 13}$.

\section{SIGNER}
In this section, we describe SIGNER, which implements time-resolved conditioning by using temporal-gloss condition and local temporal fusion (LTF) (Figure~\ref{fig:overview}).
We first introduce part-aware motion latents learned by PVQ-VAE, then construct the temporal-gloss condition, and finally describe time-resolved conditioning with LTF.

\subsection{Motion latent space learning}\label{subsec:motion_latent}
As illustrated in Figure~\ref{fig:overview}, we learn the motion latent space using PVQ-VAE to enhance the expressiveness of hand and face, inspired by prior works~\cite{soke, chen2024body_of_language, humantomato, g2p-ddm}.
Following the notations in previous section, each input motion sequence $\mathbf{X}$ is decomposed into $\mathbf{X}^\text{body}$, $\mathbf{X}^\text{lhand}, \mathbf{X}^\text{rhand}$, and $\mathbf{X}^\text{face}$.
We denote the input motion sequence of each part by $\mathbf{X}^\text{part}$, where $\text{part} \in \{\text{body}, \text{lhand}, \text{rhand}, \text{face}\}$.

The input motion sequence of each part $\mathbf{X}^\text{part}$ is processed by a dedicated 1D convolutional encoder \(E^{\text{part}}\) and decoder \(G^{\text{part}}\), with a corresponding codebook of each part \(\mathcal{Z}^{\text{part}} = \{\mathbf{z}_k^{\text{part}} \in \mathbb{R}^\text{d}\}_{k=1}^K\).  
Each input \(\mathbf{X}^{\text{part}}\) is encoded as \(\mathbf{Z}^{e, \text{part}} = E^{\text{part}}(\mathbf{X}^{\text{part}})\), then quantized using the codebook into \(\mathbf{Z}^{q, \text{part}}\).
The quantized feature is decoded as \(\hat{\mathbf{X}}^{\text{part}} = G^{\text{part}}(\mathbf{Z}^{q, \text{part}})\). 
Here, \(\mathbf{Z}^{e, \text{part}}\) and \(\mathbf{Z}^{q, \text{part}}\) denote the encoder output and the quantized latent of each part, respectively.
The training objective minimizes the loss across all parts, \(\mathcal{L} = \mathcal{L}^\text{body} + \mathcal{L}^\text{lhand} + \mathcal{L}^\text{rhand} + \mathcal{L}^\text{face}\), with each term defined in Equation~\ref{eq:vqvae_loss}.

% \begin{figure*}[t]
%     \centering
%     \includegraphics[width=\linewidth]{fig/overview4.pdf}
%     \caption{\textbf{Overview.} 
%     Our method aims to model temporal correspondence between the glosses and their associated sign segments (temporal grounding).
%     We first learn a motion latent space using PVQ-VAE (top-left).
%     Next, we construct temporal-gloss condition from the input text (bottom-left), providing time-resolved conditioning signals.
%     Finally, SIGNER generates sign language sequences by integrating temporal-gloss conditions within localized temporal windows via local temporal fusion (LTF), enabling accurate temporal grounding.
%     }
%     % \vspace{-3mm}
%     \label{fig:overview}
% \end{figure*}

\begin{figure*}[t]
    \centering
    % \vspace{-2mm}
    \begin{subfigure}[t]{0.49\linewidth}
        \centering
        \includegraphics[height=3.cm]{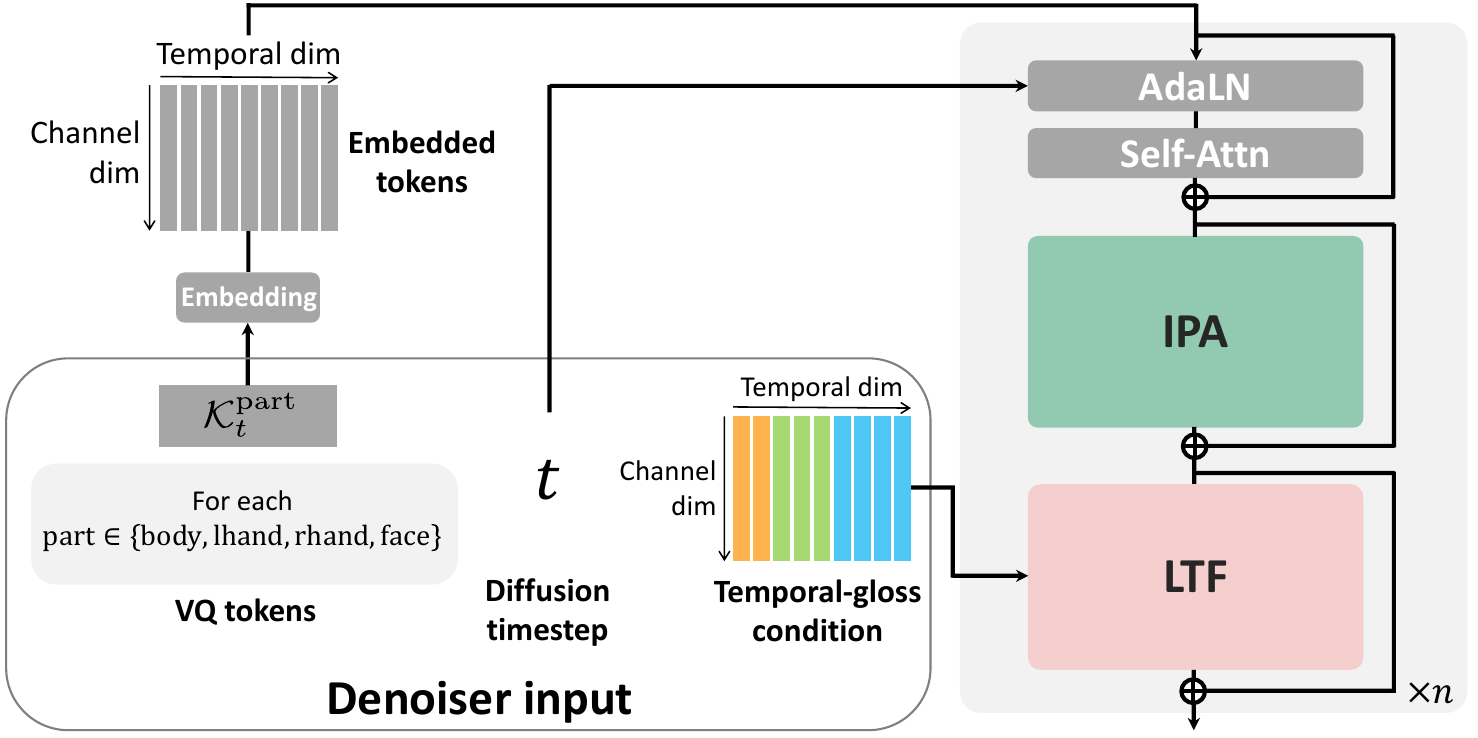}
        \caption{{Denoiser.}}
        \label{subfig:denoiser}
    \end{subfigure}
    \hfill
    \begin{subfigure}[t]{0.49\linewidth}
        \centering
        \includegraphics[height=3.cm]{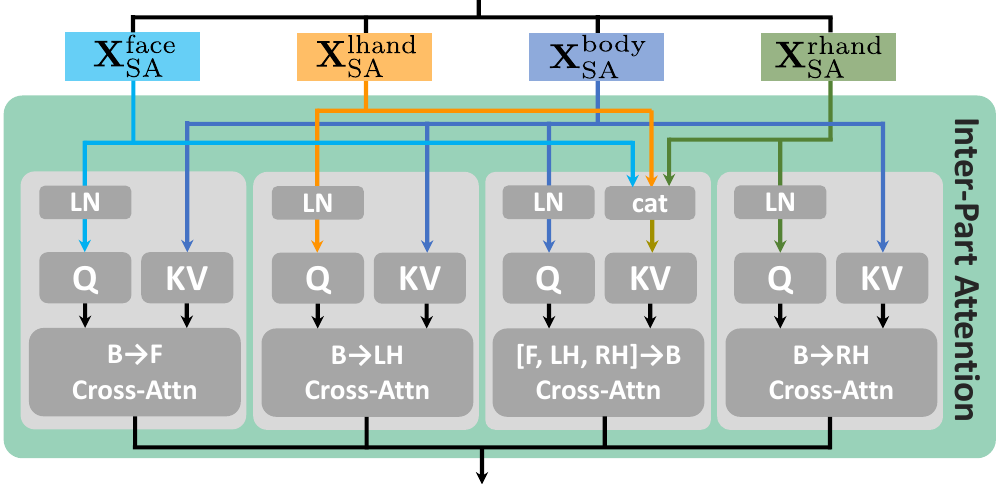}
        \caption{{Inter-Part Attention (IPA).}}
        \label{subfig:ipa}
    \end{subfigure}
    \caption{\textbf{Model architecture.} We illustrate the detailed architecture of our proposed denoiser and inter-part attention (IPA) module.
    (a) In our model, each body part—body, left hand, right hand, and face—is encoded independently without inter-part interaction, except within the IPA module.
    (b) Cross-attention between different body parts in the IPA block facilitates more coordinated movements of articulations, leading to a better quality of generated sign language.
    }
    \vspace{-3mm}
    \label{fig:denoiser}
\end{figure*}

\begin{figure*}[t]
    \centering
    \begin{subfigure}[t]{0.49\linewidth}
        \centering
        \includegraphics[height=3.4cm]{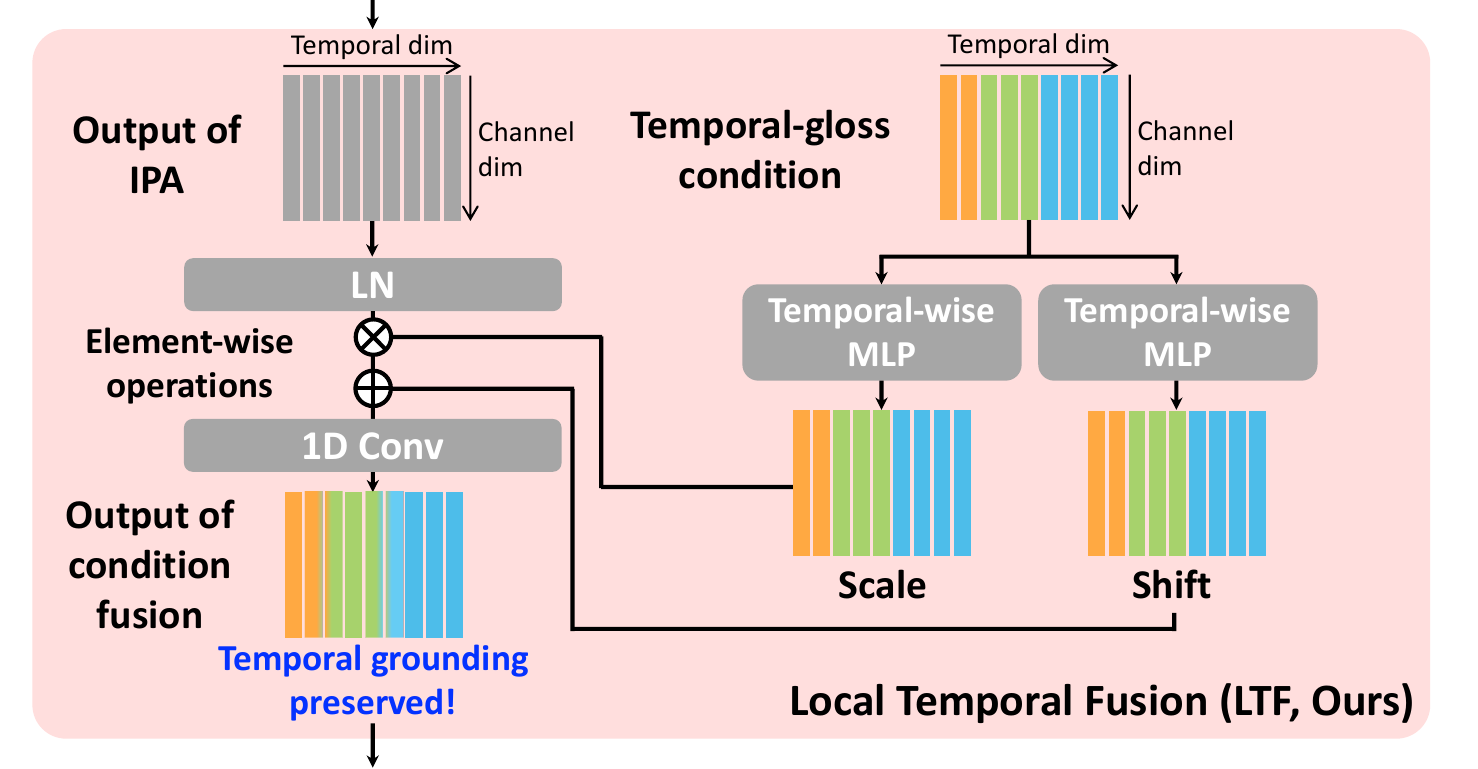}
        \caption{{Local Temporal Fusion (LTF, \textbf{Ours}).}}
        \label{subfig:adaln}
    \end{subfigure}
    \hfill
    \begin{subfigure}[t]{0.49\linewidth}
        \centering
        \includegraphics[height=3.4cm]{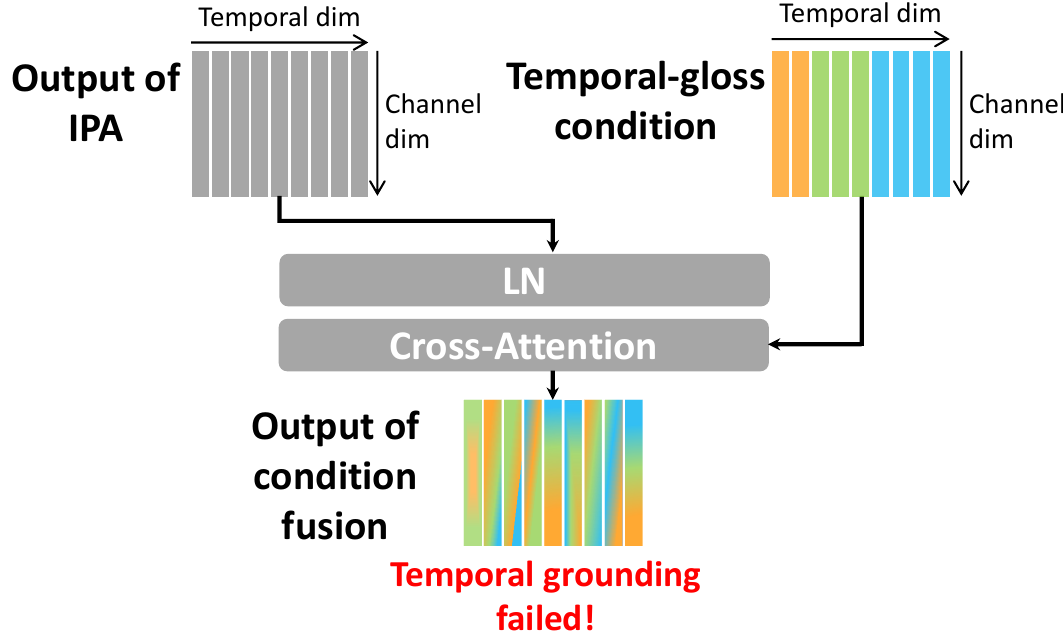}
        \caption{{Cross-attention.}}
        \label{subfig:cross_attention}
    \end{subfigure}
    \caption{\textbf{Comparison of condition fusion methods.}
    % (a) AdaLN used in ours with temporal-wise MLP and element-wise operations to keep the temporal alignment constraint.
    (a) Our LTF consists of AdaLN and 1D convolution to preserve the temporal grounding within local temporal context.
    (b) Cross-attention, which fails to maintain the temporal grounding.
    }
    \vspace{-5mm}
    \label{fig:tac}
\end{figure*}

\subsection{Temporal-gloss condition}\label{subsec:gloss_cond}
As illustrated in Figure~\ref{fig:overview} (bottom left), we first convert the input text into a gloss sequence $g=\{g_j\}_j$ using an off-the-shelf text-to-gloss model used in~\cite{spoken2sign}.
For each gloss $g_j$, we extract an embedding $c_j \in \mathbb{R}^{D_{\text{cond}}}$ using a pre-trained mBART~\cite{mbart}.
A learned duration estimator (MLP) predicts token-level durations \(l_j\) from \(c_j\), and we define the generation length as \(L_g=\sum_j l_j\).
We then construct an intermediate sequence \(\bar{S} \in \mathbb{R}^{L_g \times D_{\text{cond}}}\) by repeating each \(c_j\) for \(l_j\) steps and concatenating them in order.
Finally, we project \(\bar{S}\) to the denoiser feature dimension to obtain \(S=\Phi(\bar{S}) \in \mathbb{R}^{L_g \times D_{\text{feat}}}\).
During training and inference, we use token sequences of length $L_g$ for both the motion tokens and the temporal-gloss condition.

\subsection{Time-resolved conditioning with LTF}\label{subsec:tac}
In this section, we describe the temporally grounded sign language generation process with local temporal fusion (LTF).
First, we describe the overall architecture.
Then we introduce inter-part attention (IPA) to enable coordinated movements across articulators, which is critical when separately encoding each body part with PVQ-VAE.
We also present LTF to promote temporal localization during condition fusion.

\noindent\textbf{Architecture.}
Figure~\ref{subfig:denoiser} visualizes the architecture of the proposed denoiser.
Each body part—body, left hand, right hand, and face—is forwarded independently without inter-part interaction, except within the IPA module.
To be specific, for each part, we first represent discrete tokens $\mathbf{k}^\text{part} \in \{1,\dots,K\}^{L_g}$ as a vector of codebook indices.
The discrete tokens are formed into embedded tokens $\mathbf{X}^\text{part}_\text{embed} \in \mathbb{R}^{L_g \times D_\text{feat}}$.
$D_\text{feat}$ denotes the channel dimension of the feature, and is maintained throughout the denoiser forward pass.
Then, we integrate the diffusion timestep \( t \) into the embedded tokens $\mathbf{X}^\text{part}_\text{embed}$ using adaptive layer normalization (AdaLN)~\cite{adain}, resulting in \( \mathbf{X}^\text{part}_\text{time} \).
This is passed through a self-attention (SA) layer followed by a residual connection, yielding  
\( \mathbf{X}^\text{part}_\text{SA} = \text{SA}(\mathbf{X}^\text{part}_\text{time}) + \mathbf{X}^\text{part}_\text{embed} \).
Next, \( \mathbf{X}^\text{part}_\text{SA} \) is processed by the inter-part attention (IPA) module with a residual connection, producing  
\( \mathbf{X}^\text{part}_\text{IPA} = \text{IPA}(\mathbf{X}^\text{part}_\text{SA}) + \mathbf{X}^\text{part}_\text{SA} \).  
Finally, the features are fused with the temporal-gloss condition \( S \) in LTF, resulting in the output \( \mathbf{X}^\text{part}_\text{LTF} \).

\noindent\textbf{Inter-part attention (IPA).}
Figure~\ref{subfig:ipa} illustrates the architecture of our proposed IPA module.  
It is designed to enhance the denoising process by enabling coordinated movements across articulators through inter-part feature integration.  
Each of body features (body, hands, face) is first normalized with LayerNorm~\cite{ba2016layer} and then used as a query in cross-attention with other body parts.  
Specifically, the body features attend to a concatenation of the face and hand features, while each hand and face attends to the body features. 
The model captures dependencies between body parts through this cross-attention mechanism, producing refined features for the body and both hands.

\noindent\textbf{Local temporal fusion (LTF).}
Figure~\ref{subfig:adaln} shows LTF for temporal grounding, which integrates the temporal-gloss condition within a local temporal window.
LTF anchors the gloss semantics to their corresponding token index from the temporal-gloss condition, and aggregates nearby context to produce smooth and natural transitions.

Given a temporal-gloss condition \( S \in \mathbb{R}^{L_g \times D_\text{feat}} \), two MLPs produce per-token scale and shift, $u, v \in \mathbb{R}^{L_g \times D_\text{feat}}$.
We modulate the denoiser features with adaptive normalization and then apply a temporal 1D convolution:
\[
\mathbf{X}^\text{part}_\text{LTF} =
\text{Conv1D}\!\left(\text{LN}(\mathbf{X}^\text{part}_\text{IPA}) \odot (1 + u) + v \right),
\]
where the convolution aggregates local temporal context within a fixed window.
Through this design, AdaLN ensures that each gloss condition is aligned to its corresponding token index, while the 1D convolution allows the model to consider local context for coherent generation.
In contrast to global fusion (e.g., cross-attention over all gloss tokens), LTF confines conditioning to local neighborhoods, improving temporal grounding and transition coherence.

\noindent\textbf{Training objective.}
We follow the standard discrete diffusion objective in VQ-diffusion~\cite{vqdiffusion}.
Given a clean token sequence \(\mathbf{k}_0\) and its noised version \(\mathbf{k}_t\) at diffusion step \(t\), the denoiser predicts the posterior over the original tokens conditioned on the temporal-gloss condition $p_\theta(\mathbf{k}_0 \mid \mathbf{k}_t, t, S)$.
We train the model by minimizing the negative log-likelihood of the ground-truth tokens:
\[
\mathcal{L}_{\text{diff}} = 
\mathbb{E}_{t,\,\mathbf{k}_0,\,\mathbf{k}_t}\Big[-\log p_\theta(\mathbf{k}_0 \mid \mathbf{k}_t, t, S)\Big],
\]
where \(t\) is sampled uniformly and \(\mathbf{k}_t\) is obtained by the forward noising process.

\section{Experiment}

\subsection{Protocols}~\label{subsec:implementation_detail}

\noindent\textbf{Datasets.}
We use two large-scale sign language datasets: CSL-Daily~\cite{csldaily} and PHOENIX-2014T~\cite{phoenix}, which contain approximately 20K and 8K samples, respectively, in Chinese and German sign languages. 
Our models are trained with a maximum motion length of 180 frames~\cite{mdm, t2mgpt}, and all evaluations are conducted on the test set.

\noindent\textbf{Implementation details.}
We implement our framework using PyTorch~\cite{pytorch}. 
Input text is converted into gloss sequences via an off-the-shelf model used in~\cite{spoken2sign}, and encoded into temporal-gloss condition with mBART~\cite{mbart} embeddings.
We adopt the same hyperparameters for PVQ-VAE as used in T2M-GPT~\cite{t2mgpt}. 
Our discrete diffusion model follows the configuration of VQ-Diffusion~\cite{vqdiffusion}, except that we use 50 timesteps.
The denoiser consists of 14 layers with an inner dimension of 512 and 16 attention heads. 
LTF is implemented with 1D convolutional network of stride 1 and kernel size 3.
Gloss duration estimator consists of 2 MLP layers with 512 dimensions.
We train the models using AdamW~\cite{adamw} with a learning rate of \(1 \times 10^{-4}\) and weight decay of \(4.5 \times 10^{-2}\). 
PVQ-VAE and the denoiser are both trained for 30K iterations, with a mini-batch size of 56 on a single NVIDIA RTX 4090 GPU.

\noindent\textbf{Evaluation metrics.}
Following prior work~\cite{soke,nsa,spoken2sign}, we evaluate generated signs using metrics for motion accuracy and linguistic fidelity.
For motion-level evaluation, we use dynamic time warping with joint position error (DTW-JPE)~\cite{dtw}, which measures distances between temporally aligned joint trajectories of generated and reference motions.
For linguistic fidelity, we adopt a back-translation pipeline~\cite{chen2022twostream}: we report word error rate (WER)~\cite{wer} for sign-to-gloss, and BLEU-4~\cite{bleu} and ROUGE~\cite{rouge} for sign-to-text.

Since our method uses a new motion representation including face, evaluating our method with prior evaluators that use skeleton-only inputs can be confounded (limiting evaluation of facial expression).
Therefore, for fair comparison, we re-evaluate all methods under a unified setup: we train all models to use the same motion representation (including face as part of the input/output feature), and evaluate them with a single back-translation model trained on the same representation.
Details of the unified evaluator are provided in the supplementary material; for transparency, we additionally report cross-evaluator results using the prior evaluator in the supplementary material.

\begin{table}[t]
\footnotesize
\centering
\setlength\tabcolsep{5pt}
\def\arraystretch{1.2}
\resizebox{\textwidth}{!}{
\begin{tabular}{
    >{\centering\arraybackslash}m{5.5cm}  % Conditioning method
    >{\centering\arraybackslash}m{1.2cm}  % Body
    >{\centering\arraybackslash}m{1.2cm}  % Hand
    >{\centering\arraybackslash}m{1.4cm}  % All
    >{\centering\arraybackslash}m{1.2cm}  % WER
    >{\centering\arraybackslash}m{1.8cm}  % BLEU-4
    >{\centering\arraybackslash}m{1.6cm}  % ROUGE
}
\specialrule{.1em}{.05em}{.05em}
\hline
\multirow{2}{*}{Condition fusion} & \multicolumn{3}{c}{DTW-JPE$\downarrow$} & \multicolumn{3}{c}{Back translation} \\
\cline{2-4} \cline{5-7}
& Body & Hand & All & WER$\downarrow$ & BLEU-4$\uparrow$ & ROUGE$\uparrow$ \\
\specialrule{.15em}{.1em}{.1em}
Cross Attention & 0.251 & 1.258 & 0.832 & 84.90 & 5.81 & 24.87 \\
AdaLN + FC & 0.249 & 1.217 & 0.807 & 72.52 & 7.85 & 27.90 \\
\hline
\textbf{AdaLN + 1D Conv (LTF, Ours)} & \textbf{0.242} & \textbf{1.189} & \textbf{0.788} & \textbf{55.05} & \textbf{15.60} & \textbf{39.52} \\
\hline
\specialrule{.1em}{.05em}{.05em}
\end{tabular}}
\vspace{2mm}
\caption{\textbf{Effectiveness of LTF.}
We validate the contribution of LTF by comparing different condition fusion methods. Our method appears in the last row.}
% \vspace{-2mm}
\label{tab:ablation_condition}
\end{table}

\begin{table}[t]
\footnotesize
\centering
\setlength\tabcolsep{5pt}
\def\arraystretch{1.2}
\resizebox{\linewidth}{!}{
\begin{tabular}{
    >{\centering\arraybackslash}m{0.8cm}  % B→H
    >{\centering\arraybackslash}m{0.8cm}  % H→B
    >{\centering\arraybackslash}m{0.8cm}  % E→B
    >{\centering\arraybackslash}m{1.1cm}  % Body
    >{\centering\arraybackslash}m{1.1cm}  % Hand
    >{\centering\arraybackslash}m{1.1cm}  % All
    >{\centering\arraybackslash}m{1.2cm}  % WER
    >{\centering\arraybackslash}m{1.8cm}  % BLEU-4
    >{\centering\arraybackslash}m{1.6cm}   % ROUGE
}
\specialrule{.1em}{.05em}{.05em}
\hline
\multicolumn{3}{c}{Attention} & \multicolumn{3}{c}{DTW-JPE$\downarrow$} & \multicolumn{3}{c}{Back translation} \\
\cline{1-3} \cline{4-6} \cline{7-9}
B$\rightarrow$H & H$\rightarrow$B & B$\rightarrow$F & Body & Hand & All & WER$\downarrow$ & BLEU-4$\uparrow$ & ROUGE$\uparrow$ \\
\specialrule{.15em}{.1em}{.1em}
\xmark & \xmark & \xmark & 0.248 & 1.197 & 0.796 & 61.31 & 13.27 & 36.42 \\
\checkmark & \xmark & \xmark & 0.247 & 1.195 & 0.794 & 57.01 & 13.76 & 37.74 \\
\xmark & \checkmark & \xmark & \textbf{0.240} & 1.190 & \textbf{0.788} & 58.74 & 14.52 & 38.13 \\
\xmark & \xmark & \checkmark & 0.242 & 1.191 & 0.790 & 58.96 & 13.66 & 37.38 \\
\checkmark & \xmark & \checkmark & 0.243 & 1.194 & 0.792 & 55.31 & 15.06 & 38.95 \\
\xmark & \checkmark & \checkmark & 0.248 & 1.206 & 0.801 & 57.77 & 13.41 & 37.62 \\
\checkmark & \checkmark & \xmark & 0.247 & \textbf{1.189} & 0.790 & 57.34 & 13.59 & 37.67 \\
\textbf{\checkmark} & \textbf{\checkmark} & \textbf{\checkmark} & 0.242 & \textbf{1.189} & \textbf{0.788} & \textbf{55.05} & \textbf{15.60} & \textbf{39.52} \\
\hline
\specialrule{.1em}{.05em}{.05em}
\end{tabular}}
\vspace{2mm}
\caption{\textbf{Effectiveness of IPA.} 
We validate the contribution of IPA by testing various combinations of attention flows: Body$\rightarrow$Hands (B$\rightarrow$H), Hands$\rightarrow$Body (H$\rightarrow$B), and Body$\rightarrow$Face (B$\rightarrow$F).
Our method appears in the last row.}
\vspace{-3mm}
\label{tab:ablation_ipa}
\end{table}

\begin{table}[t]
% \vspace{-2mm}
\centering
\footnotesize
\setlength\tabcolsep{4.5pt}
\def\arraystretch{1.2}
\begin{subtable}[b]{0.48\textwidth}
\centering
\resizebox{\textwidth}{!}{
\begin{tabular}{
    >{\centering\arraybackslash}m{3cm}
    >{\centering\arraybackslash}m{0.9cm}
    >{\centering\arraybackslash}m{0.9cm}
    >{\centering\arraybackslash}m{0.9cm}
    >{\centering\arraybackslash}m{1.0cm}
    >{\centering\arraybackslash}m{1.6cm}
    >{\centering\arraybackslash}m{1.5cm}
}
\specialrule{.1em}{.05em}{.05em}
\hline
\multirow{2}{*}{\makecell{Text-to-gloss\\method}} & \multicolumn{3}{c}{DTW-JPE$\downarrow$} & \multicolumn{3}{c}{Back translation} \\
\cline{2-4} \cline{5-7}
& Body & Hand & All & WER$\downarrow$ & Bleu-4$\uparrow$ & ROUGE$\uparrow$ \\
\specialrule{.15em}{.1em}{.1em}
GT gloss     & 0.234 & 1.145 & 0.768 & 54.70 & 15.69 & 40.33 \\
\hline
mT5 finetuned     & \textbf{0.237} & \textbf{1.151} & \textbf{0.764} & 58.70 & 14.87 & 38.92 \\
M2M finetuned     & 0.250 & 1.215 & 0.806 & 67.63 & 14.79 & 39.08 \\
mBART finetuned   & 0.252 & 1.204 & 0.801 & 70.48 & 14.78 & 38.44 \\
Ours (w. off-the-shelf in~\cite{spoken2sign}) & 0.242 & 1.189 & 0.788 & \textbf{55.05} & \textbf{15.60} & \textbf{39.52} \\
\hline
\specialrule{.1em}{.05em}{.05em}
\end{tabular}
}
\caption{Robustness to text-to-gloss models.}
\end{subtable}
\hfill
\begin{subtable}[b]{0.48\textwidth}
\centering
\resizebox{\textwidth}{!}{
\begin{tabular}{
    >{\centering\arraybackslash}m{3.4cm}
    >{\centering\arraybackslash}m{0.9cm}
    >{\centering\arraybackslash}m{0.9cm}
    >{\centering\arraybackslash}m{0.9cm}
    >{\centering\arraybackslash}m{1.0cm}
    >{\centering\arraybackslash}m{1.6cm}
    >{\centering\arraybackslash}m{1.5cm}
}
\specialrule{.1em}{.05em}{.05em}
\hline
\multirow{2}{*}{Duration variation} & \multicolumn{3}{c}{DTW-JPE$\downarrow$} & \multicolumn{3}{c}{Back translation} \\
\cline{2-4} \cline{5-7}
& Body & Hand & All & WER$\downarrow$ & Bleu-4$\uparrow$ & ROUGE$\uparrow$ \\
\specialrule{.15em}{.1em}{.1em}
GT (test set) duration & 0.201 & 0.448 & 0.345 & 57.76 & 14.44 & 38.94 \\
\hline
Estimated duration + 4 & 0.247 & 1.204 & 0.799 & 56.60 & 15.18 & 38.71 \\
Estimated duration - 4 & 0.251 & 1.211 & 0.804 & 59.25 & 14.47 & 36.97 \\
Estimated duration + $t \sim U(-4, 4)$ & 0.254 & 1.219 & 0.810 & 60.74 & 14.06 & 35.53 \\
Estimated duration (Ours) & \textbf{0.242} & \textbf{1.189} & \textbf{0.788} & \textbf{55.05} & \textbf{15.60} & \textbf{39.52} \\
% length - 2 & 0.208 & 0.458 & 0.365 & 54.90 & 15.92 & 39.90 \\
\hline
\specialrule{.1em}{.05em}{.05em}
\end{tabular}
}
\caption{Robustness to gloss duration variation.}
\end{subtable}
\caption{\textbf{Robustness of our system.} 
We show the robustness of our method on the choice of text-to-gloss models and the variation of gloss duration.
Our method appears in the last row.
}
\vspace{-3mm}
\label{tab:ablation_robustness}
\end{table}

\begin{table*}[t]
\footnotesize
\centering
\setlength\tabcolsep{4.0pt}
\def\arraystretch{1.2}
\resizebox{\textwidth}{!}{
\begin{tabular}{
    >{\centering\arraybackslash}m{1.2cm}  % Setting
    >{\centering\arraybackslash}m{3cm}    % Method
    >{\centering\arraybackslash}m{1.0cm}  % CSL - Body
    >{\centering\arraybackslash}m{1.0cm}  % CSL - Hand
    >{\centering\arraybackslash}m{1.0cm}  % CSL - All
    >{\centering\arraybackslash}m{1.3cm}  % CSL - WER
    >{\centering\arraybackslash}m{1.5cm}  % CSL - BLEU-4
    >{\centering\arraybackslash}m{1.3cm}  % CSL - ROUGE
    >{\centering\arraybackslash}m{0.3cm}  % divider
    >{\centering\arraybackslash}m{1.0cm}  % PHX - Body
    >{\centering\arraybackslash}m{1.0cm}  % PHX - Hand
    >{\centering\arraybackslash}m{1.0cm}  % PHX - All
    >{\centering\arraybackslash}m{1.3cm}  % PHX - WER
    >{\centering\arraybackslash}m{1.5cm}  % PHX - BLEU-4
    >{\centering\arraybackslash}m{1.3cm}  % PHX - ROUGE
}
\specialrule{.1em}{.05em}{.05em}
\hline
\multirow{3}{*}{Setting}
& \multirow{3}{*}{Method}
& \multicolumn{6}{c}{\textbf{CSL-Daily}} & & \multicolumn{6}{c}{\textbf{Phoenix-2014T}} \\
\cline{3-8} \cline{10-15}
&
& \multicolumn{3}{c}{DTW-JPE$\downarrow$} & \multicolumn{3}{c}{Back translation} &
& \multicolumn{3}{c}{DTW-JPE$\downarrow$} & \multicolumn{3}{c}{Back translation} \\
\cline{3-8} \cline{10-15}
& 
& Body & Hand & All & WER$\downarrow$ & BLEU-4$\uparrow$ & ROUGE$\uparrow$ 
& & Body & Hand & All & WER$\downarrow$ & BLEU-4$\uparrow$ & ROUGE$\uparrow$ \\
\specialrule{.15em}{.1em}{.1em}

\multirow{5}{*}{\shortstack{Gloss\\free}}
& MotionGPT~\cite{motiongpt}
& 0.939 & 2.258 & 1.700 & 94.92 & 1.79 & 15.48
&& 1.709 & 3.424 & 2.699 & 99.60 & 0.81 & 7.07 \\

& MDM~\cite{mdm}
& 0.351 & 1.754 & 1.161 & 95.54 & 2.78 & 16.29
&& 0.855 & 2.819 & 1.988 & 99.83 & 0.97 & 7.40 \\

& NSA~\cite{nsa}
& 0.347 & 1.691 & 1.123 & 96.64 & 2.09 & 14.56
&& 0.826 & 2.406 & 1.738 & 98.29 & 1.03 & 7.62 \\

& MoMask~\cite{momask}
& 0.376 & 1.565 & 1.062 & 91.51 & 3.79 & 20.12
&& 0.864 & 2.296 & 1.690 & 92.01 & 5.15 & 22.80 \\

& SOKE~\cite{soke}
& 0.256 & 1.283 & 0.849 & 85.69 & 4.09 & 21.70
&& 0.231 & 1.102 & 0.733 & 87.74 & 5.52 & 19.69 \\

\hline

\multirow{3}{*}{\shortstack{Gloss\\based}}
& G2P-DDM~\cite{g2p-ddm}
& 0.249 & 1.248 & 0.825 & 74.28 & 6.44 & 25.31
&& 0.236 & 1.138 & 0.757 & 88.00 & 5.17 & 16.78 \\

& NAT-EA~\cite{natat}
& 0.532 & 1.598 & 1.147 & 67.80 & 7.87 & 29.17
&& 0.453 & 1.475 & 1.043 & 75.65 & 8.79 & 23.04 \\

& \textbf{Ours}
& \textbf{0.242} & \textbf{1.189} & \textbf{0.788} & \textbf{55.05} & \textbf{15.60} & \textbf{39.52}
&& \textbf{0.229} & \textbf{1.096} & \textbf{0.729} & \textbf{67.16} & \textbf{11.46} & \textbf{29.39} \\

\hline
\specialrule{.1em}{.05em}{.05em}
\end{tabular}
}
\vspace{2mm}
\caption{\textbf{Comparison to state-of-the-art methods.}
We report motion errors and back-translation results under a unified evaluation setting on the CSL-Daily and Phoenix-2014T datasets.
Methods are grouped into gloss-free and gloss-based settings for clearer comparison.
Our SIGNER outperforms previous methods on both datasets.
}
\vspace{-5mm}
\label{tab:main_csl_phoenix}
\end{table*}

\subsection{Ablation Study}

\noindent\textbf{Local temporal fusion (LTF).}
Table~\ref{tab:ablation_condition} shows that time-resolved conditioning with LTF yields substantial performance gains.
Global fusion via cross-attention (row 1) yields limited generation performance, as it allows each motion timestep to aggregate information from all gloss tokens, which can blur temporal grounding.
In contrast, our LTF (row 3) restricts conditioning to a local temporal window, and thereby preserves temporal grounding. 
This localized fusion leads to accurate signs in the correct lexical order.
Replacing the temporal convolution with a fully connected layer (row 2) consistently degrades performance, confirming that local temporal aggregation is critical for incorporating nearby context and producing smooth transitions.
Overall, these results indicate that time-resolved conditioning in diffusion process effectively addresses the limitations of global fusion used in prior methods.

\noindent\textbf{Inter-part attention (IPA).}
Table~\ref{tab:ablation_ipa} shows that IPA improves sign language generation performance notably.
As illustrated in Figure~\ref{subfig:ipa}, IPA integrates inter-part dependencies through cross-attention flows between body and hands (B$\rightarrow$H and H$\rightarrow$B) as well as body-to-face attention (B$\rightarrow$F).
Individually enabling B$\rightarrow$H or H$\rightarrow$B attention (second and third rows) yields consistent gains over the baseline without IPA (first row), highlighting the importance of coordinating body and hand articulations.
Incorporating B$\rightarrow$F attention further improves performance (fourth and fifth rows), indicating that facial cues provide complementary information beyond body--hand interactions.
Notably, enabling all attention flows (B$\rightarrow$H, H$\rightarrow$B, and B$\rightarrow$F; last row) achieves the best overall results.
Based on these findings, we adopt this full IPA configuration as our final design.

\noindent\textbf{Robustness on text-to-gloss and duration sampling.}
Table~\ref{tab:ablation_robustness} evaluates the robustness of our framework to variations in both the text-to-gloss model and the predicted sign duration.
The left block reports results using alternative text-to-gloss models (fine-tuned mT5~\cite{mt5}, M2M~\cite{m2m}, and mBART~\cite{mbart}).
Performance remains strong across different text-to-gloss choices, indicating that SIGNER is not overly sensitive to the specific text-to-gloss model.

The right block (Table~\ref{tab:ablation_robustness} (b)) tests robustness to perturbations around the estimated sign duration.
We test robustness by perturbing the estimated duration for each gloss by +4, -4, and a random offset sampled from $U(-4,4)$.
Across these settings, performance remains stable, suggesting that SIGNER is not overly sensitive to moderate errors or noise in the predicted duration.
Given that the average gloss duration in CSL-Daily is 16 frames, these perturbations correspond to approximately $\pm 25\%$ variation, supporting the temporal robustness of our method.

\subsection{Comparison to state-of-the-art methods}\label{exp:comparison_sota}
Table~\ref{tab:main_csl_phoenix} demonstrates that our proposed SIGNER substantially outperforms previous methods on both Phoenix-2014T~\cite{phoenix} and CSL-Daily~\cite{csldaily}.

\begin{figure}[t]
\centering
\begin{minipage}[t]{0.49\linewidth}
  \centering
  \footnotesize
  \setlength\tabcolsep{5pt}
  \def\arraystretch{1.2}
  \vspace{-10mm}
  \resizebox{0.95\linewidth}{!}{
  \begin{tabular}{
      >{\centering\arraybackslash}m{3cm}
      >{\centering\arraybackslash}m{3cm}
      >{\centering\arraybackslash}m{3cm}
  }
  \specialrule{.1em}{.05em}{.05em}
  \hline
  Model & Peak velocity ($\downarrow$) & Peak jerk ($\downarrow$) \\
  \specialrule{.15em}{.1em}{.1em}
  NSA~\cite{nsa}    & 0.75$\pm$0.20 & 1.75$\pm$0.59 \\
  MoMask~\cite{momask} & 0.44$\pm$0.13 & 0.63$\pm$0.31 \\
  Spoken2Sign~\cite{spoken2sign}    & 0.88$\pm$1.11 & 1.70$\pm$2.14 \\
  NAT-EA~\cite{natat} & 0.43$\pm$0.10 & 0.69$\pm$0.31 \\
  \textbf{Ours}   & \textbf{0.38}$\pm$\textbf{0.10} & \textbf{0.35}$\pm$\textbf{0.12} \\
  \hline
  \specialrule{.1em}{.05em}{.05em}
  \end{tabular}}
  \vspace{10mm}
  \captionof{table}{\textbf{Peak jerk and peak velocity.}
  We report peak velocity and peak jerk which indicates the naturalness and smoothness of generated motion.
  Results are reported in normalized units.}
  \label{tab:peak_jerk}
\end{minipage}\hfill
\begin{minipage}[t]{0.45\linewidth}
  \centering
  \includegraphics[width=\linewidth, height=2cm]{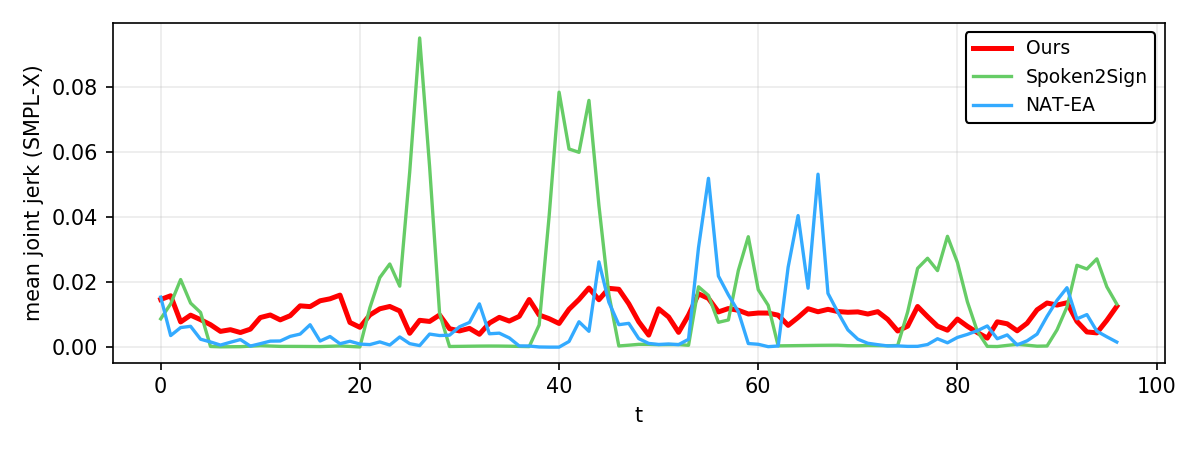}\\[-1mm]
  \includegraphics[width=\linewidth, height=2cm]{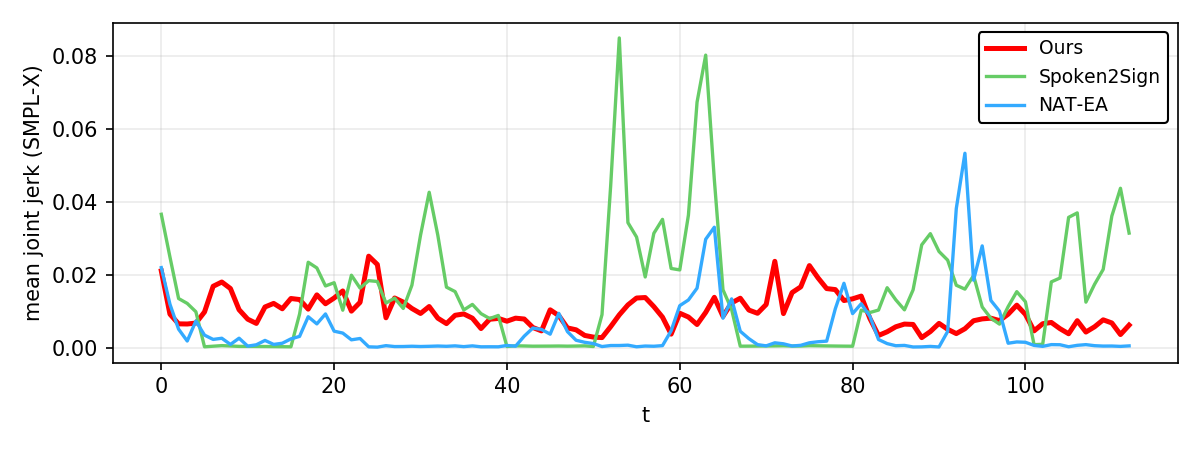}
  \vspace{-3mm}
  \caption{\textbf{Jerk curves.}
  We visualize mean joint jerk over time for two samples. 
  Lower peak height indicates smoother transitions.}
  \label{fig:jerk_curves}
  
\end{minipage}
\vspace{-2mm}
\end{figure}

\begin{figure}[t!]
  \centering
  \begin{subfigure}{\linewidth}
    \centering
    \includegraphics[width=\linewidth]{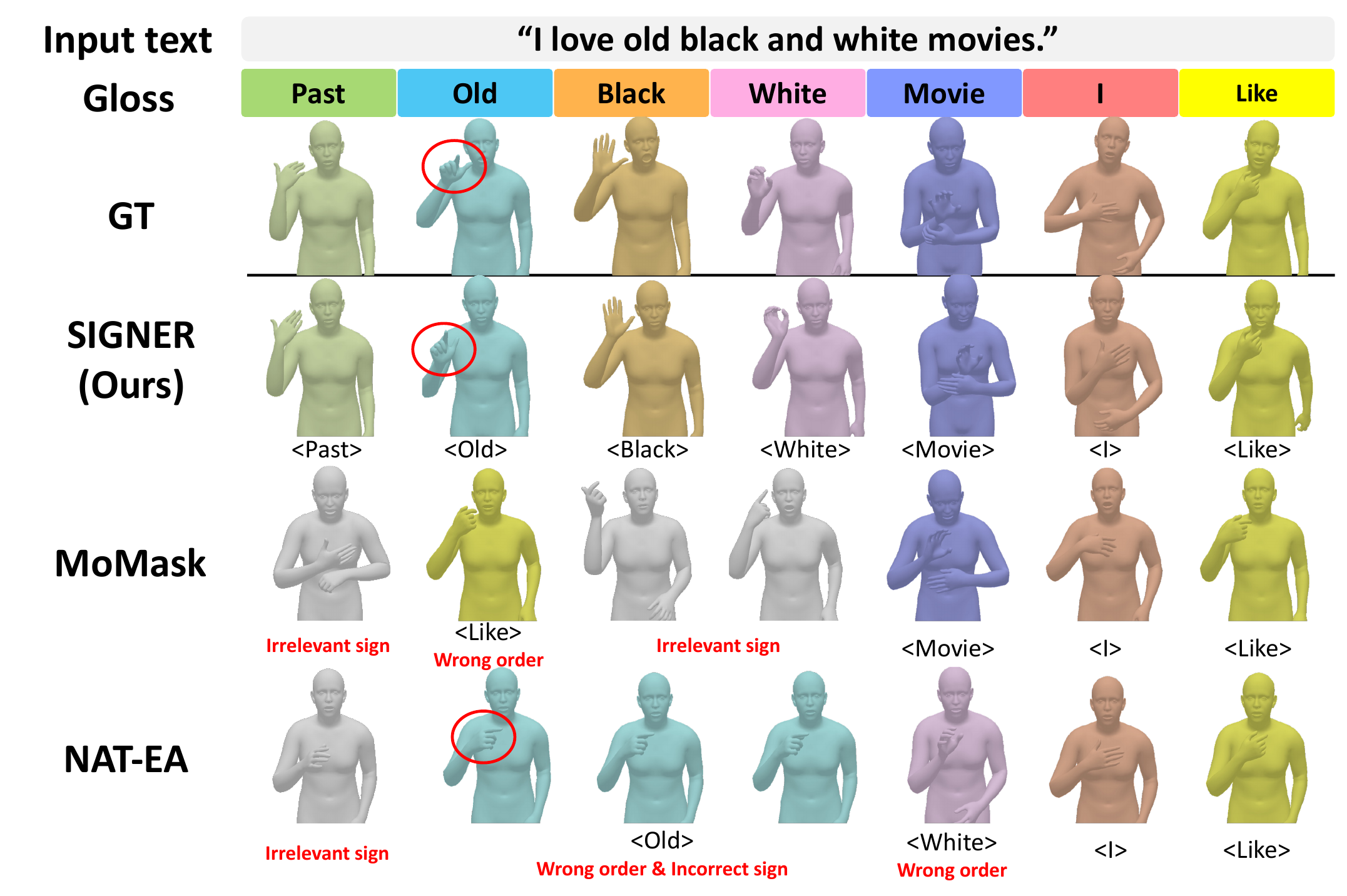}
  \end{subfigure}
  \begin{subfigure}{\linewidth}
    \centering
    \includegraphics[width=\linewidth]{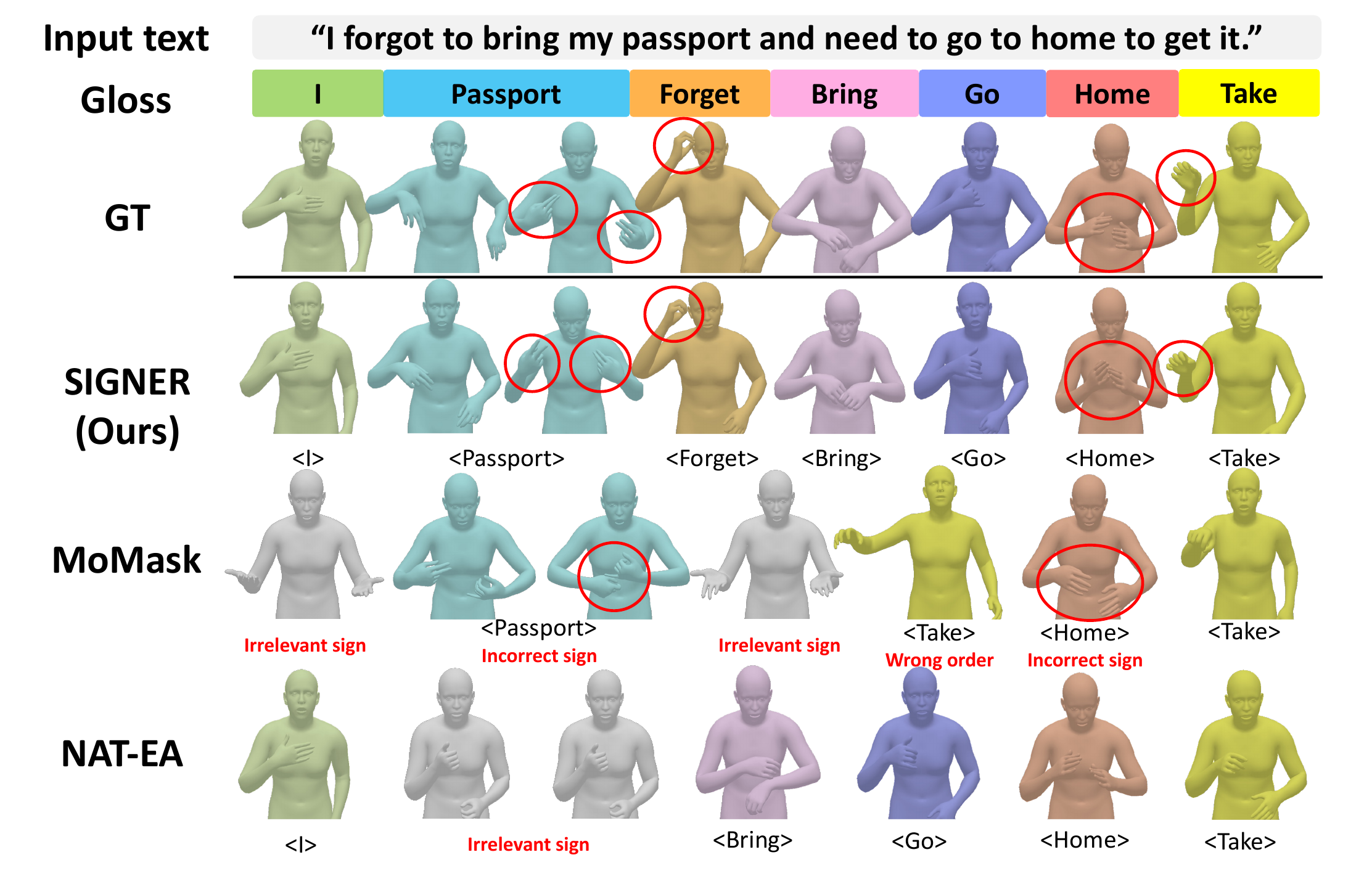}
  \end{subfigure}
  \caption{\textbf{Qualitative result.} We present the qualitative comparison of our SIGNER with GT, MoMask~\cite{momask}, and NAT-EA~\cite{natat}, all trained on CSL-Daily.}
  \label{fig:qualitative}
\end{figure}

A key takeaway is that how conditions are fused across time is critical.
On CSL-Daily (left block), SIGNER achieves state-of-the-art performance among all prior methods. 
Compared to global-fusion approaches such as MoMask~\cite{momask} and SOKE~\cite{soke}, SIGNER outperforms by a large margin. 
Methods that incorporate gloss-level cues, including G2P-DDM~\cite{g2p-ddm} and NAT-EA~\cite{natat}, generally outperform text-only baselines, yet SIGNER still yields substantially higher scores. 
Notably, the performance gap between our SIGNER and NAT-EA~\cite{natat}, which also leverages a temporal-gloss condition, suggests that using temporal-gloss information alone is insufficient to guarantee correct lexical ordering and semantic accuracy.
Rather, the conditioning mechanism must preserve temporal grounding during generation.
Similarly, on Phoenix-2014T (right block), SIGNER delivers consistent improvements over existing methods across all evaluation metrics. 
These gains across datasets further support that time-resolved conditioning by LTF enhances sign generation quality by ensuring temporal grounding.

\noindent\textbf{Motion smoothness and transition quality.}
We evaluate motion smoothness and transition quality using peak velocity and peak jerk, where lower values indicate smoother motions.
As shown in Table~\ref{tab:peak_jerk}, SIGNER achieves the lowest peak velocity and peak jerk among the compared methods.
Figure~\ref{fig:jerk_curves} provides qualitative evidence: our jerk curves exhibit lower peaks, whereas baselines show pronounced peaks around transitions.
We include Spoken2Sign~\cite{spoken2sign} as a representative retrieval-and-concatenation baseline to contextualize transition quality.

\noindent\textbf{Qualitative result.}
Figure~\ref{fig:qualitative} shows that our SIGNER generates accurate sign language in correct lexical order compared with MoMask~\cite{momask} and NAT-EA~\cite{natat}.
Each gloss is represented by its own colors, and the corresponding sign segment is shown in the same color.
Human-interpreted glosses are denoted inside brackets.
Compared with our method, MoMask conditions the denoiser on a text feature via global fusion, which does not preserve temporal grounding; as a result, it often produces signs with incorrect lexical order and semantically mismatched segments.
NAT-EA leverages a temporal-gloss condition, but it still fuses the condition globally, so temporal grounding is only partially preserved.
As a result, NAT-EA still generates signs with incorrect lexical order and semantically mismatched segments.

\section{Conclusion}
% We propose SYNC, a sign language generation framework with temporally localized conditioning.
% By incorporating temporal-gloss conditions and TAC, SYNC addresses key limitations of previous conditioning methods: the lack of temporal structure and granularity to convey fine-grained context of each timestep.
% By overcoming two limitations above, our SYNC successfully generates accurate sign language.
% Our proposed method highly outperforms previous methods on CSL-Daily and Phoenix-2014T.
\noindent\textbf{Summary.}
We presented SIGNER, a temporally grounded sign language generation framework that addresses two central limitations of prior SLG methods: incorrect lexical ordering and low semantic accuracy.
We identified globally fused conditioning as a key cause, as it weakens temporal grounding—the temporal correspondence between glosses and their realized sign segments—by mixing linguistic cues across time.
To preserve temporal grounding during generation, SIGNER implements time-resolved conditioning with a temporal-gloss condition and local temporal fusion (LTF), which injects gloss semantics within localized temporal windows during diffusion denoising.
Experiments on Phoenix-2014T and CSL-Daily demonstrate that SIGNER achieves state-of-the-art performance under a unified evaluation setup, showing that our proposed time-resolved conditioning efficiently addresses the challenge of lexical ordering and semantic accuracy.

\noindent\textbf{Limitation.}
While SIGNER demonstrates strong performance in sign language generation, it relies on off-the-shelf text-to-gloss models.
Despite this dependency, Table~\ref{tab:ablation_robustness}(a) shows that SIGNER remains robust across different text-to-gloss models. 
However, to train SIGNER with a new sign language dataset on a different language system, a text-to-gloss model must still be trained to extract the necessary gloss annotations. 
Future work could explore jointly optimizing text-to-gloss and sign generation within a unified framework to alleviate this dependency.

\section*{Acknowledgements}
This work was supported in part by the IITPgrants [No. RS-2021-II211343, Artificial Intelligence Graduate School Program (Seoul National University), No. RS-2025-02303870, No.2022-0-00156] funded by the Korea government (MSIT).

\bibliographystyle{splncs04}
\bibliography{main}
\end{document}